\documentclass[sigconf,natbib=true,nonacm]{acmart}
\renewcommand\footnotetextcopyrightpermission[1]{} 
\usepackage{enumitem}
\usepackage{multirow}
\usepackage{subfigure}
\usepackage{caption}
\usepackage{graphicx}
\usepackage{bbm}
\usepackage{array}
\usepackage{algorithmic}
\usepackage{caption}

\usepackage[linesnumbered, ruled]{algorithm2e}
\newcommand{{\method}}{DisenLink}
\newcommand{{\methoda}}{DisenLink-$\alpha$}
\newcommand{{\methods}}{DisenLink-s}
\newcommand{{\methodr}}{DisenLink-r}
\newcolumntype{P}[1]{>{\centering\arraybackslash}p{#1}}

\usepackage{booktabs}
\usepackage{hyperref}
\hypersetup{hidelinks,
	colorlinks=true,
	allcolors=black,
	pdfstartview=Fit,
	breaklinks=true}

\newtheorem{problem}{Problem}
\AtBeginDocument{%
  \providecommand\BibTeX{{%
    \normalfont B\kern-0.5em{\scshape i\kern-0.25em b}\kern-0.8em\TeX}}}

\setcopyright{none}
\copyrightyear{}
\acmYear{}
\acmDOI{}

\begin{document}

\title{Link Prediction on Heterophilic Graphs via Disentangled Representation Learning}

\author{Shijie Zhou$^\dagger$, Zhimeng Guo$^\dagger$, Charu Aggarwal$^\ddagger$, Xiang Zhang$^\dagger$, Suhang Wang$^\dagger$ }

\affiliation{{\textsuperscript{\textdagger}}College of Information Sciences and Technology, The Pennsylvania State University, USA\\
{$^{\ddagger}$}IBM T. J. Watson Research Center
 \country{USA}
 }

\email{{smz5479,zzg5107,xzz89,szw494}@psu.edu, charu@us.ibm.com}


\begin{abstract}
Link prediction is an important task that has wide applications in various domains. However, the majority of existing link prediction approaches assume the given graph follows homophily assumption, and designs similarity-based heuristics or representation learning approaches to predict links. However, many real-world graphs are heterophilic graphs, where the homophily assumption does not hold, which challenges existing link prediction methods. Generally, in heterophilic graphs, there are many latent factors causing the link formation, and two linked nodes tend to be similar in one or two factors but might be dissimilar in other factors, leading to low overall similarity. Thus, one way is to learn disentangled representation for each node with each vector capturing the latent representation of a node on one factor, which paves a way to model the link formation in heterophilic graphs, resulting in better node representation learning and link prediction performance. However, the work on this is rather limited. Therefore, in this paper, we study a novel problem of exploring disentangled representation learning for link prediction on heterophilic graphs. We propose a novel framework {\method} which can learn disentangled representations by modeling the link formation and perform factor-aware message-passing to facilitate link prediction. Extensive experiments on 13 real-world datasets demonstrate the effectiveness of {\method} for link prediction on both heterophilic and hemophiliac graphs. Our codes are available at \url{https://github.com/sjz5202/DisenLink}.

\end{abstract}



\maketitle

\section{Introduction}
Graph structured data are pervasive in real-world, such as social networks~\cite{yang2018graph}, knowledge graphs~\cite{Nickel2016ARO} and transaction networks~\cite{ceylan2021learning}. For many domains, the collected graph usually contains many unknown or missing links due to various reasons such as limitation of observation capacity and artificial masking. For example, the constructed knowledge graph has many missing relationships due to the incompleteness of knowledge. 
Thus, link prediction, which aims to predict missing links in a graph, is an active research area. Various link prediction algorithms have been proposed~\cite{barabasi2002evolution,koren2009matrix,zhang2017weisfeiler}, which have facilitate various applications such as friend suggestion in social networks~\cite{yang2018graph}, knowledge graph completion~\cite{Nickel2016ARO} and protein-protein interaction reconstruction ~\cite{lei2013novel}. Existing link prediction based approaches can be generally categorized into two categories, i.e., heuristic-based approaches~\cite{newman2001clustering,adamic2003friends,jeh2002simrank} and representation learning based approaches~\cite{acar2009link,kipf2016variational}. Heuristic-based approaches measure local or global similarity within a specific graph topology or node features based on the corresponding assumption to give the probability of the edge existence. For example, SimRank~\cite{jeh2002simrank} gives the score based on the similarity between 1-hop neighbors of two nodes. Representation learning methods learn node representations that can well reconstruct the links. For example, matrix factorization-based link prediction~\cite{acar2009link} optimize the coefficient matrix as final representations. 
Recently, graph neural network (GNN) based representation learning approaches~\cite{zhang2018link} for link prediction have achieved state-of-the-art performance. Generally, GNNs adopt the message-passing mechanism, which iterative aggregates neighbor nodes' representations to update the center node's representation. Thus, the learned representation captures both node attributes and local graph information, which facilitates link prediction.

\begin{figure}[t]
\centering
    \includegraphics[width=0.6\linewidth]{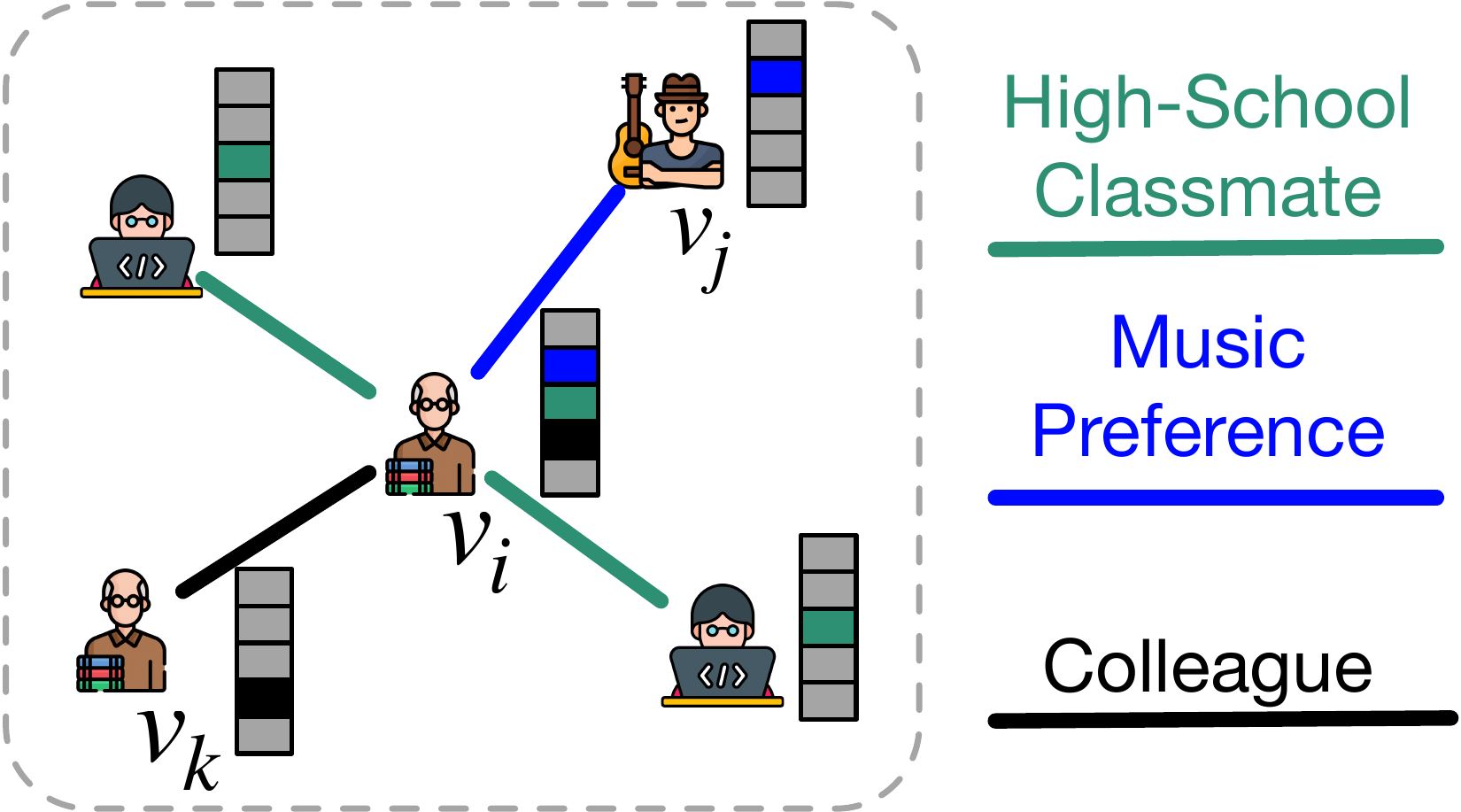}
    \caption{Social network with different relationships.}     \label{fig:social}
\end{figure}

However, the aforementioned approaches generally assume that the given graph follows homophily assumption, i.e., two nodes with similar features are more likely to be connected; while many real-world graphs are heterophilic, where a node tends to connect to nodes of dissimilar features or class labels due to various complex latent factors rather than the simple homophily assumption. For example, as shown in Figure~\ref{fig:social},  in social networks, two users $v_i$ and $v_j$ might be connected as they have common interests in music but have no similarity in other factors; $v_i$ and $v_k$ might be connected simply because they are colleagues but have totally different interests and cultural background. Similarly, in molecular graphs, many latent factors influence the bonds between atoms. Often the case, the reason why two nodes are connected is unknown. Hence, \textit{in heterophilic graphs, two connected nodes might be similar in one or two latent factors but have low overall feature similarity}, which breaks the homophily assumption and challenges many existing link prediction approaches, especially GNN based approaches~\cite{kipf2016variational,zhang2018link} as (1) Generally, GNN adopts message-passing to learn node representation, which aggregates neighborhood features to update a node's representation. Directly performing message-passing on heterophilic graphs could aggregate neighbors of dissimilar features of the center nodes, causing noisy representations, which degrades the performance of downstream tasks~\cite{rong2019dropedge,zhang2018end}; (2) Representation learning based approach simply learn node representation and adopt homophily assumption to make predictions, i.e., a pair of nodes with similar representations are more likely to be linked, which might not be true in heterophilic graphs. Thus, link prediction on heterophilic graphs is a challenging problem. However, the work in this direction is rather limited.

As there are many latent factors that affecting the formation of links in heterophilic graphs and two nodes are connected due to one or two latent factors, one natural way to perform link prediction is to learn $K$ disentangled representations $\left\{\mathbf{h}_{i,k}\right\}_{k=1}^{K}$ with $\mathbf{h}_{i,k}$ for each node $i$ with $\mathbf{h}_{i,k}$ capturing the characteristics of $v_i$ on factor $K$. With the disentangled representation, we can model the formation of links between each pair of nodes, i.e., which factor is the one that most likely causes the formation of the link. It has several advantages: (i) we can perform message-passing on important factors while ignoring unimportant factors to avoid aggregating noisy information. Hence, we can learn better disentangled representations for link prediction; (ii) we do not rely on homophily assumption as we can perform link prediction with disentangled representations. Even though the overall representation of two nodes is small, if they share a common interest in some factors, we would still be able to give the right prediction. There are several existing disentangled learning methods on graphs. However, they are not directly applicable for link prediction on heterophilic graphs as (i) they aim to learn disentangled representation for node~\cite{ma2019disentangled} or graph~\cite{yang2020factorizable} classification; and (ii) they do not model the link formation and are not dedicated to heterophilic graphs. Thus, though it is a promising direction, there are no existing works on exploring disentangled representation for link prediction on heterophilic graphs. 

Therefore, in this paper, we study a novel problem of exploring disentangled representation for link prediction on heterophilic graphs. i.e. learning disentangled node representations from GNNs for link prediction in heterophilic graphs.  There are several challenges for link prediction on heterophilic graphs: (i)  How to learn high-quality disentangled representations that capture the most informative and pertinent latent factor without the supervision of latent factor selection? (ii) How to model link formation of representations from multiple factors? (iii) How can we learn the disentangled representations that capture the neighborhood information beyond the noisy aggregation in real-world heterophilic graphs for better link predictions? To address these challenges, in this work, we propose a link prediction model named {\method}. {\method} focuses on edge factor discovery at the beginning to select the most important edge factor for each potential connectivity based on the similarity between local representations for informative factors. {\method} then takes the relational similarity information and graph locality into consideration and performs factor-wise message passing within the neighbor of the selected factor to produce better disentangled representations. {\method} does not naively reconstruct links following VGAE~\cite{kipf2016variational} but considers different contributions of each factor for better link prediction. The main contributions of the paper are:
\begin{itemize}[leftmargin=*]
\item We study a novel problem of disentangled representation learning for link prediction on heterophilic graphs;
\item We propose a novel framework that designs factor-wise message passing with factor-aware neighbor selection to learn high-quality disentangled representation for link prediction;
\item We conduct experiments on both heterophilic and homophilic graphs to demonstrate the effectiveness of {\method}.
\end{itemize}

\section{Related Works}

\subsection{Link Prediction}
Link prediction, which aims to predict if two nodes in a network are likely to have a link, is an important task. It has wide applications in various domains such as social networks and knowledge graphs. Thus, various approaches have been proposed, which can be generally categorized into two categories~\cite{kumar2020link,lu2011link}, i.e., heuristics-based approaches~\cite{adamic2003friends, katz1953new} and representation learning based approaches~\cite{kipf2016variational, zhang2018link}. 
Heuristics-based approach computes the pair-wise similarity score based on a particular structure or node's properties. For example, common neighbors (CN)~\cite{newman2001clustering} is scored by the number of the neighborhood intersection between a pair of nodes. Adamic Adar (AA)~\cite{adamic2003friends} is measured based on the assumption that the unpopular common neighbor is more likely to formulate the connection between two nodes. These heuristics-based approach tends to make strong assumptions and fails to generalize to different graph structures. More seriously, they cannot effectively utilize the attribute information, which severely limits the performance of the model. 
Representation learning based approaches aim to learn good node representations that can well reconstruct the adjacency matrix.  They usually apply dot product between two node representations to predict the likelihood of the link between two nodes. 
For example, matrix factorization-based link prediction~\cite{acar2009link,menon2011link} adopts matrix factorization to learn node representations for adjacency matrix reconstruction. Due to the great ability in learning node representations that can capture both node and graph structure information,  GNNs have become state-of-the-art representation learning based approaches for link prediction~\citep{zhang2018link,kipf2016variational}. For example, VGAE ~\citep{kipf2016variational} adopts the GNN as the encoder to learn node representations followed by a simple inner product decoder to get the link prediction result. SEAL extracts an enclosing subgraph centered at two nodes followed by a GNN on the graph to predict if there is a link between the two nodes~\cite{zhang2018link}. 

Despite their great success, most of them assume homophily assumption, i.e., two nodes with similar neighbors, features, or representations are more likely to form a link. Their capacity on heterophilic graphs is doubtful as the homophily assumption on heterophilic graphs does not hold. In heterophilic graphs, linked nodes tend to have dissimilar features and different labels. For example, in dating networks, most people tend to build a connection with people of the opposite gender~\citep{zhu2020graph}; in transaction networks, fraudsters prefer to connect with customers instead of other fraudsters~\citep{pandit2007netprobe}. Thus,  link prediction for heterophilic graphs needs further investigation. However, the work in this direction is rather limited.

Therefore, in this work, we study a novel problem of link prediction on heterophilic graphs. We propose a novel framework to learn disentangled representation by modeling the underlying diverse and complex connectivity relation in heterophilic graphs and to use the disentangled representation to predict missing links beyond the homophily assumption.

\subsection{Graph Neural Networks}
Graph Neural Networks (GNNs) have shown great power in representations on graph-structured data. GNNs can be generally categorized into two categories, spectral-based and spatial-based~\cite{hamilton2020graph}. Spectral-based GNNs such as GCN~\cite{kipf2016semi} and ChebNet~\cite{defferrard2016convolutional} utilize the convolutional operator in the spectral domain of the Laplacian matrix. Spatial-based GNNs such as GraphSAGE~\cite{hamilton2017inductive}, LGCN~\cite{gao2018large} and GAT~\cite{velivckovic2017graph} perform message passing directly on the graph topology. The success of GNNs relies on the message-passing mechanism, i.e., a node will iteratively aggregate the information from its neighbors. Thus, the learned representation captures both node attributes and local graph structure information, which facilitate various downstream tasks such as node classification~\cite{rong2019dropedge}, graph classification~\cite{zhang2018end}, and link prediction~\cite{zhang2018link}.

Though GNNs have achieved great success in various domains, the message-passing mechanism of GNNs implicitly assumes the homophily assumption of the graph~\citep{zhu2020beyond}, i.e., nodes with similar features or same class labels are linked together; however, not all real-world graphs obey the homophily assumption~\citep{zheng2022graph}. Simply applying message passing on heterophilic graphs would learn noisy node representation as a node's neighbors can have very dissimilar features with the node.  
Thus, how to extend GNNs for heterophilic graphs has attracted increasing attention and many efforts have been take~\cite{Pei2020Geom-GCN,chien2020adaptive,lim2021new,zhu2020beyond}. For example, Geom-GCN~\cite{Pei2020Geom-GCN} adds structural neighbors in continuous space to capture long-range dependencies for heterophilic graphs. 

To model the intrinsic factors in graph-structure data, there are some works~\cite{yang2020factorizable,ma2019disentangled,liu2020independence} that concentrate on GNNs with a disentangled model.
DisenGCN~\citep{ma2019disentangled} is an early-stage work and its motivation is to 
utilize a neighborhood routing mechanism to find latent factors that shape the relationship between the nodes. For each factor, the nodes aggregate the information and get different representations.
FactorGNN employs different MLP to get edge scores and several factor graphs. They also consider an adversarial training scheme to discriminate different factors.
Besides these works, \cite{zhao2022exploring} comes up with a new approach that effectively utilizes self-supervision to constrain the disentangled representations to capture distinguishable relations and neighborhood interactions.

Our work is inherently different from the aforementioned GNNs for heterophilic graphs or disentanglement based GNNs: (i) They focus on node or graph classification task, which has labels to guide the learning; while we study a novel problem of link prediction in heterophilic graphs; (ii) Though they also try to learn disentangled representation, we develop novel factor-aware neighbor selection and factor-wise message passing to model the link formation and learn disentangled node representation, and propose the novel disentangled link reconstruction for the final link formation.

\section{Problem Definition}
We use $\mathcal{G}=(\mathcal{V}, \mathcal{E}, \mathbf{X})$ to denote an attributed graph, where $\mathcal{V}=\{v_1, ..., v_N\}$ is the set of $N$ nodes, $\mathcal{E}$ is the set of observed edges, and $\mathbf{X}=\{\mathbf{x}_1,...,\mathbf{x}_N\}$ node feature matrix with $\mathbf{x}_i$ being the attributes of node $v_i$. $\mathbf{A} \in \mathbb{R}^{N \times N}$ is the adjacency matrix of $\mathcal{G}$, where $\mathbf{A}_{s t}=1$ if nodes $v_s$ and $v_t$ are connected; otherwise $\mathbf{A}_{st}=0$. Many real-world graphs such as social networks and knowledge graphs usually have many missing/unseen links. The task of link prediction is to infer the missing links. Most existing link prediction algorithms~\cite{kipf2016variational,zhang2017weisfeiler,zhang2018link} assume that the given graph follows homophily assumption, i.e., two nodes of the same class or similar node features are more likely to be connected. 
However, many real-world graphs are heterophilic graphs, where the homophily assumption does not hold. In a heterophilic graph, a node tends to connect to nodes of dissimilar features or class labels due to various complex latent factors rather than the simple homophily assumption. For example,  in social networks, two users $v_i$ and $v_j$ might be connected as they have common interests in music but have no similarity in other factors; $v_i$ and $v_k$ might be connected simply because they are colleagues but have totally different interests and cultural background. 

Hence, connected nodes in heterophilic graphs might be similar in one or two factors but have low overall feature similarity, 
which challenges existing link prediction approaches with homophily assumption, especially state-of-the-art GNN based approaches~\cite{kipf2016variational,zhang2018link} as (1) For heterophilic graphs, the direct performing of message-passing in the general GNN scheme will aggregate neighbors of dissimilar features of the center nodes leading noisy representations, which degrades the performance of downstream tasks~\cite{rong2019dropedge,zhang2018end}; (2) Representation learning based approach simply learn node representation and adopt homophily assumption to make predictions, i.e., two nodes with similar representation are more likely to be connected.

As there are many latent factors that affecting the formation of links and two nodes are connected due to one or two latent factors, for each node $v_i$, we propose to learn $K$ disentangled representation $\left\{\mathbf{h}_{i,k}\right\}_{k=1}^{K}$ with $\mathbf{h}_{i,k}$ capturing the characteristics of $v_i$ on factor $k$. Then for a pair of nodes $(v_i,v_j)$, we can utilize disentangled representation to model which factor is most likely for $v_i$ and $v_j$ to be connected, which can simultaneously facilitate message-passing under the most important factor to avoid aggregating noise and help factor-wise link prediction to avoid simple similarity based link prediction. The problem of disentangled representation learning on heterophilic graphs for link prediction is formally defined as
\vspace*{-0.5em}
\begin{problem}
Given a heterophilic graph $\mathcal{G}=(\mathcal{V}, \mathcal{E}, \mathbf{X})$ whose links are formed due to various latent factors. Note that the ground truth of the factor that causes the formation of each link is unknown.  Learn disentangled representation of each node for link prediction on $\mathcal{G}$.
\end{problem}

\begin{figure*} 
    \centering
    \includegraphics[width=0.86\linewidth]{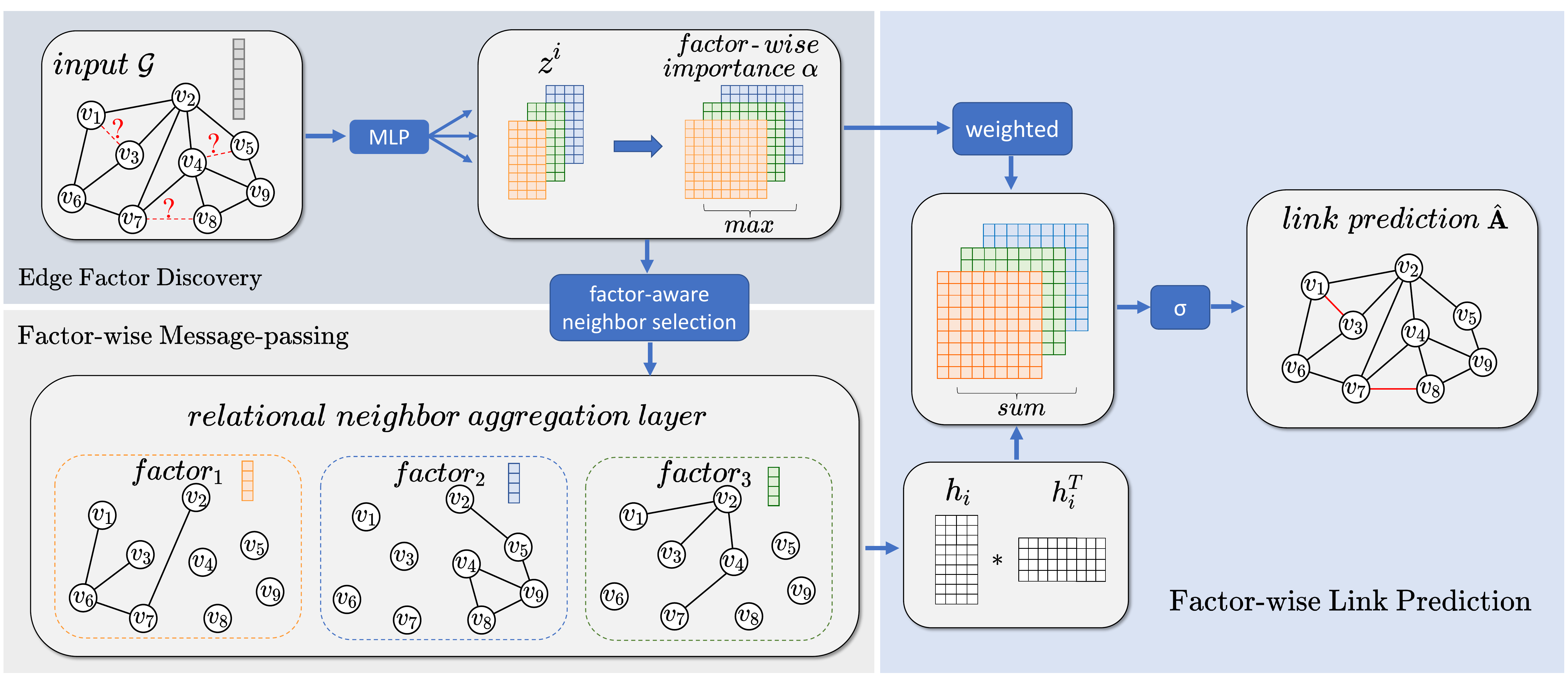}
    \vskip -1.5em
    \caption{An illustration of the proposed framework {\method}} \label{fig:model}
    \vskip -1em
\end{figure*}

\section{METHODOLOGY}
In this section, we will give the details of  {\method} to learn disentangled node representations for link prediction on heterophilic graphs. As there are various factors underlying the graph formation and a node is connected to another node due to one or two factors, the basic idea is to learn $K$ disentangled representation vectors $\{\mathbf{z}_{s,k}\}_{k=1}^K$ for each node $v_s$ with the $i$-th vector $\mathbf{z}_{s,i}$ capturing $v_s$'s representation on the $i$-th factor, and utilize the disentangled representations to model the link formation. However, there are several challenges: (i)  How to model link formation and learn high-quality disentangled representations that capture latent factor information of each node given that we do not have the supervision of which latent factor contributes most to the formation of a link? (ii) Since directly applying message-passing mechanism can aggregate noisy information from neighbors, how can we learn disentangled representations that capture the neighborhood information for better link prediction? To solve these challenges, we propose a novel framework {\method} as shown in Fig.~\ref{fig:model}, which is composed of Factor-Aware Neighbor Selection, Factor-Wise Message-Passing, and Link Prediction with Disentangled Representation. For each edge $(v_i,v_j)$, factor-aware neighbor selection considers all factors and designs a mechanism to find which factor is the most likely factor that forms the edge $(v_i,v_j)$.  Factor-wise message-passing only performs message-passing on the most likely factor to avoid aggregating noisy information from neighbors. Hence, {\method} can learn better disentangled representation that captures heterophilic neighborhood information. With disentangled representations, {\method} finally adopts factor-wise link prediction to predict the missing link.  Next, we introduce details of each component.

\subsection{Factor-Aware Neighbor Selection}
Generally, a node's raw feature $\mathbf{x}$ contains all the information on various factors, but is entangled. To help learn $K$ disentangled representation, for each node $v_s$, we first project the raw feature $\mathbf{x}_s$ to $K$ feature spaces to get the initial disentangled representations. Specifically, the representation of $v_s$ on the $k$-th factor is given as 
\begin{equation}
    \mathbf{z}_{s,k}=f_k(\mathbf{x}_{s}), \quad k=1, \dots, K
\end{equation}
where $f_k$ is the $k$-th project function. Various projection functions can be used for $f_k$. In this paper, we choose a two-layer MLP, i.e., 
\begin{equation} \label{eq:mlp}
\mathbf{z}_{s,k}=\mathbf{W}_k^{(2)} \cdot ReLU(\mathbf{W}_k^{(1)}\mathbf{x}_{s})
\end{equation}
where $\mathbf{W}_k^{(1)}$ and $\mathbf{W}_k^{(2)}$ are the parameters of the first layer and second layer MLP for factor $k$, respectively.

The initial disentangled representation $[\mathbf{z}_{s,1}, ..., \mathbf{z}_{s,K}]$ only disentangled the node feature information. It does not take the local graph structure and neighbor features into consideration; while various studies~\citep{liu2021local} have demonstrated the importance of local graph information for link prediction. Message-passing mechanism of GNNs has shown great ability in learning representations that capture both node features and local graph information. The basic idea of message-passing is to iteratively aggregate the neighbor's representations to update a node's representation. 
However, in heterophilic graphs,  node $v_s$ and node $v_t \in {\mathcal{N}(v_s)}$ are linked due to one or two factors and could have totally dissimilar features in other factors, where ${\mathcal{N}(v_s)}$ is set of neighbors of node $v_s$. Directly applying message passing on all the $K$ disentangled representations would aggregate noisy information for most of the factors, resulting in noisy node representations. To avoid aggregating the noisy information from neighbors, for each neighbor node $v_t \in \mathcal{N}(v_s)$, we should perform message passing on one or two factors that cause the connection of $v_s$ and $v_t$. However, we lack the ground truth of these connections. Essentially, two nodes are connected because they share similarities in one or two factors. Thus, instead of relying on the ground truth, we can measure the similarity of $\mathbf{z}_{s,k}$ and $\mathbf{z}_{t,k}$, $k=1,\dots,K$, select the most similar factor, say $i$-th factor, and perform message-passing on $i$-th factor. The advantages are: (\textbf{i}) the most similar factor is likely to be the factor that causes $v_s$ and $v_t$ to be connected; and (\textbf{ii}) we avoid aggregating neighbor representations on dissimilar factors. 
Specifically, for $v_t \in {\mathcal{N}(v_s)}$, we calculate the factor-wise importance/similarity between $v_s$ and $v_t$ on the $k$-th factor as:
\begin{equation} \label{eq:importance_normed}
\alpha_{s, t}^{k}=\frac{\exp \left(\mathbf{z}_{s, k}^{T} \cdot \mathbf{z}_{t, k} / \tau\right)}{\sum_{i=1}^{K} \exp \left(\mathbf{z}_{s, i}^{T} \cdot \mathbf{z}_{t, i} / \tau\right)}, \quad {k=1,\dots,K}
\end{equation}
where $\tau$ denotes the temperature parameter. $\alpha_{s, t}^{k}$ is the the factor-wise importance score between $v_s$ and $v_t$ on factor $k$, which is basically the normalized similarity between $\mathbf{z}_{s, k}$ and $\mathbf{z}_{t, k}$ in the $k$-th factor. It can be treated as how well the factor $k$ can be the explanation of the mutual relation between $v_s$ and $v_t$.  A larger $\alpha_{s, t}^{k}$ means that $v_s$ and $v_t$ are more similar in terms of factor $k$ and it is more likely that factor $k$ is the desired factor. Since we are interested in having one element among $\{\alpha_{s,t}^k\}_{k=1}^K$ to be large while the others to be small, i.e., modeling that two nodes are connected due to only one factor, a smaller $\tau$ can be set to intensify the uneven degree of each factor-wise importance distribution. 

With $\alpha_{s, t}^{k}$ denoting the importance of the factor $k$ for the link between $v_s$ and $v_t$, we can find the most important factor $p_{s, t}$ between $v_s$ and $v_t$ as follows
\begin{equation} \label{eq:max_selection}
p_{s, t}=\arg \max _{k} \alpha_{s, t}^{k}
\end{equation}
Let $\mathcal{N}_{k}(v_s)$, $k=1,\dots,K$, denote the set of $v_s$'s neighbors connected to $v_s$ due to factor $k$. Since  $p_{s, t}$ gives the most important factor between $v_s$ and $v_t$, we can obtain $\mathcal{N}_{k}(v_s)$ as
\begin{equation}
\mathcal{N}_{k}(v_s)=\left\{v_{t}: v_{t} \in \mathcal{N}(v_s) \wedge p_{s, t}=k\right\}, \quad k=1,\dots,K
\label{eq:neighbor}
\end{equation}

With the above equations, we can model the latent factors that forms the links and disentangle each node $v_s$'s neighbor set $\mathcal{N}(v_s)$ into $K$ disjoint sets, i.e., $\mathcal{N}_s = \cup_{k=1}^K \mathcal{N}_{k}(v_s)$ and $\mathcal{N}_{k}(v_s) \cap \mathcal{N}_{j}(v_s) =\emptyset$, which would facilitate the factor-wise message passing to learn better disentangled representations to be introduced next.

\subsection{Factor-Aware Message Passing}
As mentioned previously, we want to perform message passing to learn disentangled representations that capture the neighborhood information for better link prediction. Instead of simply aggregating all the neighbors' information for each factor as other GNNs do, we adopt factor-aware message passing because a node $v_s$ might be only similar with the central node in one or two factors in heterophilic graphs. Specifically, as $\mathcal{N}_{k}(v_s)$ contains neighbors of $v_s$ who share similarity with $v_s$ in $k$-th factor, $v_s$ will only aggregate the $k$-th representation from nodes in $\mathcal{N}_{k}(v_s)$. Since the importance or similarity of nodes in $\mathcal{N}_{k}(v_s)$ with $v_s$ in terms of factor $k$ could be different, to take this into consideration, we adopt the attention mechanism, i.e., assign different attention weight to different nodes in $\mathcal{N}_k(v_s)$. ,The attention score of $v_t \in \mathcal{N}_{k}(v_s)$ on $v_s$ is given as
\begin{equation} \label{eq:attention}
\bar{\alpha}_{s, t}^{k}=\frac{\alpha_{s, t}^{k}}{\sum_{v_i \in \mathcal{N}_{k}(v_s)} \alpha_{s, i}^{k}}
\end{equation}
where $\bar{\alpha}_{s,t}^k$ denotes attention score of $v_t \in \mathcal{N}_{k}(v_s)$ on $v_s$. And $\alpha_{s,t}^k$ is  from Eq.(\ref{eq:importance_normed}), which denotes the importance score or similarity between node $v_s$ and $v_t$. Eq.(\ref{eq:attention}) normalizes the importance score by the summation of the scores from nodes in $\mathcal{N}_{s,k}$. 
Thus, if $\alpha_{s,t}^k$ is larger, node $v_t$ will play a more important role in message-passing to $v_s$ for factor $k$. With the attention score, the message-passing on factor $k$ for node $v_s$ can be written as
\begin{equation}
\mathbf{h}_{s, k}=\beta \cdot \mathbf{z}_{s, k}+(1-\beta) \sum_{v_t \in \mathcal{N}_{k}(v_s)} \bar{\alpha}_{s, t}^{k} \cdot \mathbf{z}_{t, k}, \quad, k=1,\dots,K
\label{eq:layer}
\end{equation}
where $\beta\in(0,1]$ is to control the reservation of the information from central nodes. 
With the factor-aware message passing, for each node $v_s$, we obtained the new disentangled representation set $\{\mathbf{h}_{s,1},\dots, \mathbf{h}_{s,K}\}$. Compared with the raw disentangled representation $\{\mathbf{z}_{s,1},\dots, \mathbf{z}_{s,K}\}$, the new disentangled representation also captures the neighborhood information, which would better facilitate link prediction. In practice, we can stack multiple such factor-aware neighbor selection and factor-wise message passing to have deep disentangled representations. We leave the deep disentangled representation as future work.

\subsection{Link Prediction with Disentangled Representation}
Given the node representation,  a popular way to predict the link between two nodes $v_s$ and $v_t$ is using the dot product between their node representations $\mathbf{h}_s$ and $\mathbf{h}_t$, i.e., $\hat{\mathbf{A}}_{st} = \sigma(\mathbf{h}_s^T \mathbf{h}_t)$, where $\sigma$ is an activation function such as ReLU. In our case, $\mathbf{h}_s = [\mathbf{h}_{s,1}, \dots, \mathbf{h}_{s,K}]$ and $\mathbf{h}_t = [\mathbf{h}_{t,1}, \dots, \mathbf{h}_{t,K}]$. Thus,  the link can be predicted as
\begin{equation} 
    \hat{\mathbf{A}}_{st} = \sigma(\mathbf{h}_s^T \mathbf{h}_t) = \sigma\Big(\sum_{k=1}^K \mathbf{h}_{s,k}^T \mathbf{h}_{t,k}\Big)
\label{eq:link_prediciton_equally}
\end{equation}
However, the above equation treats each factor equally; while in heterophilic graph, two connected nodes could be similar in few factors and be dissimilar in most factors. Those dissimilar factors might each have negative similarity in terms of $\mathbf{h}_{s,k}^T \mathbf{h}_{t,k}$.  Simply adding those non-important factors would result in poor link prediction performance. Thus, to model how links are formed, instead of using Eq.(\ref{eq:link_prediciton_equally}),  we propose the disentangled link prediction. The basic idea is to assign different weights to different factors during link prediction. However, we do not have ground truth of which factors are more important. Fortunately, we have calculated the importance score of factor $k$ between $v_s$ and $v_t$ using the similarity between $\mathbf{z}_{s,k}$ and $\mathbf{z}_{t,k}$ in Eq.(~\ref{eq:importance_normed}). In other words, the model is trained to have $\mathbf{z}_{s,k}$ and $\mathbf{z}_{t,k}$ to give good importance similarity on factor $k$. Thus, we adopt a similar way to calculate the weight for link prediction on factor $k$ as 
\begin{equation}
\gamma_{s, t}^{k}=\exp \left(\mathbf{z}_{s, k}^{T} \mathbf{z}_{t, k} / \tau\right)
\end{equation}
Note that we used the unnormalized version compared with Eq.~(~\ref{eq:importance_normed}) because the scale of each element in $\gamma$ is useful information for link reconstruction as it preserve the contrast between the similarities of different pairs of nodes with different scales. Then the link between $v_s$ and $v_t$ is predicted as
\begin{equation}
\hat{\mathbf{A}}_{s t}=\sigma\left(\sum_{k=1}^{K}\left(\gamma_{s, t}^{k} \mathbf{h}_{s, k}^{T} \mathbf{h}_{t, k}\right)\right)
\label{eq:re}
\end{equation}
where the contribution of each factor for link prediction is weighted by $\gamma_{s,t}^k$.  Those non-important factors are down-weighted to have neglectful effect on the link prediction, which can handle the non-homophily assumption in heterophilic graphs. 

\subsection{Final Objective Function}
Following existing work~\cite{kipf2016variational}, we adopt adjacency matrix reconstruction as the loss to train the model.  Since the majority of node pairs are unconnected, most of the elements in $\mathbf{A}$ are 0. To avoid the imbalanced elements dominating the loss function, following~\cite{zhang2018link}, we adopt negative sampling to solve this issue. Specifically, we treat each linked pair as a positive sample. For each positive sample, i.e.,  $\mathbf{A}_{st}=1$, we randomly sample $M$ unlinked pairs as negative samples, i.e., $\mathbf{A}_{sm}=0$. $M$ is usually set as 5 or 10. With positive and negative samples, the overall loss function for the disentangled link reconstruction in our model can be written as
\begin{equation}
\min _{\theta} \mathcal{L}_{\mathbf{A}}=\sum_{\mathbf{A}_{s t}>0}\Big[(\hat{\mathbf{A}}_{s t}-\mathbf{A}_{s t})^{2}+\frac{1}{M} \sum_{\mathbf{A}_{s m} \in \mathcal{M}_{s t}^{-}}(\hat{\mathbf{A}}_{s m}-\mathbf{A}_{s m})^{2}\Big]
\label{eq:loss}
\end{equation}
where $\theta$ is the set of parameters. $\mathcal{M}_{st}^{-}$ is the set of negative samples for $\mathbf{A}_{st}>0$.

The training algorithm of {\method} is shown in Algorithm~\ref{alg:Framwork}. The  parameter of the {\method} is randomly initialized following ~\cite{glorot2010understanding} in line 1. Then we perform each component of {\method} from line 3 to line 7 for the final link reconstruction loss $\mathcal{L}_{\mathbf{A}}$ in line 8. The size of negative edges for $\mathcal{L}_{\mathbf{A}}$ is 5 times the positive edges to leverage the imbalanced number of linked and unconnected edges. Adam~\cite{kingma2014adam} is used to update $\theta$ until reaching convergence.


\begin{algorithm}[t]
\caption{ Training Algorithm of {\method}}
\label{alg:Framwork}
\begin{algorithmic}[1]
\REQUIRE
$\mathcal{G}=(\mathbf{V},\mathbf{A}, {X})$, $\beta$, $\tau$ and $\mathcal{M}^{-}$
\ENSURE $\theta$ of function $\left\{\mathbf{f}_{k}\right\}_{k=1}^{K}$
\STATE Randomly initialize the parameters of $\{{f}_{k}\}_{k=1}^{K}$.
\WHILE{Not Converged}
\STATE Get the initial node representations $\mathbf{Z}$ according to Eq.~(\ref{eq:mlp});
\STATE Calculating the factor-wise importance score with Eq.~(\ref{eq:importance_normed});
\STATE Implement the neighbor selection for each factor;
\STATE Propagate $\mathbf{Z}$ via the relational neighbor aggregation Layer for embedding $\mathbf{H}$ according to Eq.~[\ref{eq:layer}];
\STATE Perform the disentangled link reconstruction for reconstructed $\hat{\mathbf{A}}$ following Eq.~[\ref{eq:re}];
\STATE Calculate the link reconstruction loss $\mathcal{L}_{\mathbf{A}}$ with Eq.~[\ref{eq:loss}];
\STATE Update $\theta$ to minimize $\mathcal{L}_{\mathbf{A}}$
\ENDWHILE
\RETURN $\theta$ 
\end{algorithmic}
\end{algorithm}

\section{EXPERIMENTS}
In this section, we evaluate the effectiveness of the proposed {\method} for link prediction on both homophilic and heterophilic graphs. Further analysis is provided to evaluate the influence of components in the proposed module. Specifically, we want to answer the following questions:
\begin{itemize}[leftmargin=*]
    \item \textbf{RQ1} Can the proposed {\method}  with disentangled link reconstruction achieve outperformed accuracy for link prediction?
    \item \textbf{RQ2} How does each component in {\method} influence the effectiveness of link prediction?
    \item \textbf{RQ3} Does the {\method} with factor-wise neighborhood selection produce independent and distinguishable factors?
\end{itemize}

\subsection{Experiment Settings}
\subsubsection{Datasets} We conduct experiments on 13 publicly available benchmark datasets to evaluate the effectiveness of {\method} for link prediction.   
Among the datasets, 3 of them are homophily graphs, i.e., Cora, CiteSeer and Photo, and the remaining 10 are heterophilic graphs. These graphs are from various domains such as citation, Wikipedia and Amazon goods. We include both heterophilic and homophilic graphs of various domains to fully understand the effectiveness of {\method} for link prediction. The key statistics of the datasets are shown in Table~\ref{tab:stat}. The details of the datasets are: 
\begin{itemize}[leftmargin=*]
    \item \textbf{Cora} and \textbf{CiteSeer}~\cite{yang2016revisiting}: Both Cora and CiteSeer  are homophilic citation networks, where nodes represent documents and edges denote the citation relationships. The attributes of nodes are the bag-of-words representations of the papers. 
    \item \textbf{Photo}\cite{shchur2018pitfalls}: The Amazon Photo network is a homophilic goods network. Nodes denote goods and two connected nodes mean that they are frequently bought together.  Node features are bag-of-words representations of goods' reviews. 
    \item \textbf{Wisconsin and Texas}~\cite{Pei2020Geom-GCN}: Wisconsin and Texas are heterophilic webpage networks collected from CMU, where nodes denote webpages and edges are the hyperlinks connecting them. Node features are the bag-of-words representation of web pages. 
    \item \textbf{Chameleon, Squirrel and Crocodile}~\cite{Pei2020Geom-GCN}: Chameleon, Squirrel and Crocodile are heterophily Wikipedia webpage networks with each network on the topic denoted by its dataset name. Nodes are web pages and edges are hyperlinks between them. Each node's features denote several informative nouns on Wikipedia pages. 
    \item \textbf{Facebook100}~\cite{traud2012social,lim2021new}: Facebook100 is composed of 100 Facebook friendship networks of American universities from 2005, where nodes represent students and edges denote friendship between students. Each node is labeled with gender. 
    The node features are educational information of students. 
    We include two of them: \textbf{Johns Hopkins55} and \textbf{Amherst41} for our experiments.
    \item \textbf{Twitch-explicit}~\cite{lim2021new}: Twitch-explicit contains 7 networks where Twitch users are nodes and friendships between them are edges. Node features are games liked, location and streaming habits. We use the two largest networks: \textbf{DE} and \textbf{ENGB} for our experiments. 
    \item \textbf{arXiv-Year}~\cite{hu2020open}: arXiv-year is the ogbn-arXiv citation network with labels denoted by publication year. The nodes are arXiv papers and directed edges connect a paper to other papers that it cites. It contains a single graph with 169343 nodes and 1166243 links. The feature dimension is 128. We sample a subgraph with around 55K nodes and 153K edges utilizing the Random Walk sampler.
\end{itemize}

\begin{table}[t]
\caption{The statistics of datasets}
\vskip -1em
\resizebox{1\linewidth}{!}{
\begin{tabular}{cccccc}
\hline
Dataset & Nodes &  Edges &  Features & Classes  & Edge hom. \\ \hline
Cora & 2,708 & 5,278 & 1,433 & 7 & 0.81 \\
CiteSeer & 3,327 & 4,552 & 3,703 & 6  & 0.74 \\
Photo & 7,650 & 238,162 & 745 & 8  & 0.82\\\hline
Chameleon & 2,277 & 36,101 & 2,325 & 5 & 0.23 \\ 
Squirrel & 5,201 & 216,933 & 2,089 & 5  & 0.22 \\
Crocodile & 11,631 & 180,020 & 500 & 6  &  0.25\\
Texas & 183 & 309 & 1,703 & 5  & 0.11 \\
Wisconsin & 251 & 499 & 1,703 & 5  & 0.20 \\
Twitch-ENGB & 7,126 & 36,324 & 2,545 & 2  & 0.55  \\
Twitch-DE & 9,498 & 153,138 & 2,514 & 2  & 0.63 \\
Johns Hopkins55 & 5,180 & 373,172 & 2406 & 2  & 0.50  \\
Amherst41 & 2,235 & 181,908 & 1,193 & 2  & 0.46 \\
Arxiv-Year* & 55,104 & 153,565 & 128 & 5 & 0.22 \\ \hline
\end{tabular}
}
\label{tab:stat}
\vskip -1em
\end{table}

\begin{table*}[h]
    \centering
    \small
    \caption{Link Prediction performance (AUC(\%) $\pm$ Std.) on heterophilic graphs (``-'' denotes out of memory.)} \label{tab:heter}
    \vskip -1em
    \resizebox{1\linewidth}{!}{
    \begin{tabular}{lccccccccccc}
    \toprule
    Method & Chameleon &Squirrel & Crocodile & Texas & Wisconsin & Twitch-DE & Twitch-ENGB& Amherst41 & Johns Hopkins55 & Arxiv-Year \\
    \midrule
        AA & 95.8 $\pm 0.6$& 97.1 $\pm 0.4$& 90.6 $\pm 1.1$& 53.1 $\pm 6.2$& 61.5 $\pm 0.8$& 87.3 $\pm 0.2$&74.5 $\pm 0.4$&94.5 $\pm 0.7$&96.1 $\pm 0.5$&72.0 $\pm 0.9$\\
        CN & 95.3 $\pm 0.6$& 96.8 $\pm 0.4$& 89.8 $\pm 1.2$& 53.0 $\pm 6.2$& 61.3 $\pm 0.9$& 86.3 $\pm 0.1$&74.3 $\pm 0.2$&94.2 $\pm 0.6$&95.8 $\pm 0.6$&72.0 $\pm 0.9$\\
        SEAL&\textbf{99.5}$\pm0.1$&-&-&73.9 $\pm 1.6$&72.3 $\pm 2.7$&-& 91.5 $\pm 0.3$ &-&-&93.2$\pm0.8$&\\
        VGAE & 98.5 $\pm 0.1$ & 98.2 $\pm 0.1$ & 98.8 $\pm 0.1$ &68.6 $\pm 4.2$&71.3 $\pm 4.6$&84.3 $\pm 0.9$ &90.3 $\pm 0.2$& 97.2 $\pm 0.0$ & 92.8 $\pm 0.2$ &92.5 $\pm 0.5$  \\
        ARGVA & 96.5 $\pm 0.2$ & 93.6 $\pm 0.3$ & 94.1 $\pm 0.5$ &67.4 $\pm 6.1$&67.6 $\pm 3.0$&80.9 $\pm 1.1$&82.2 $\pm 0.8$& 81.2 $\pm 0.3$ &81.0 $\pm 0.7$  &86.8 $\pm 1.1$ \\
        GPR-GNN & 98.7 $\pm 0.1$ & 96.0 $\pm 0.3$ & 96.7 $\pm 0.1$ &76.3 $\pm 2.5$&80.1 $\pm 4.5$& 88.7 $\pm 0.1$&87.3 $\pm 0.1$& 93.7 $\pm 0.0$ & 94.7 $\pm 0.1$ & 93.0 $\pm 0.4$ \\
        FAGCN & 93.8 $\pm 2.6$ & 94.8 $\pm 0.3$ & 95.3 $\pm 0.2$ &68.7 $\pm 7.3$&73.7 $\pm 4.9$&85.4 $\pm 0.4$&89.6 $\pm 0.4$& 91.4 $\pm 0.3$  & 92.2 $\pm 0.3$  &84.8 $\pm 2.3$ \\
        LINKX & 98.8 $\pm 0.1$ & 98.1 $\pm 0.3$ & 99.0 $\pm 0.1$ &75.8 $\pm 4.7$&80.1 $\pm 3.8$&88.3 $\pm 1.9$&89.3 $\pm 1.6$& 93.5 $\pm 0.2$& 93.4 $\pm 0.3$ &88.4  $\pm 1.9$ \\
        GAT & 98.9 $\pm 0.1$ & 98.0 $\pm 0.0$ & 98.5 $\pm 0.1$ &68.5 $\pm 5.4$&68.1 $\pm 4.4$&87.4 $\pm 0.4$&86.1 $\pm 1.9$& 93.9 $\pm 0.1$ & 94.3 $\pm 0.5$ &92.5 $\pm 0.5$\\
        DisenGCN & 97.7 $\pm 0.1$ & 94.5 $\pm 0.2$ & 96.4 $\pm 0.4$ &72.1 $\pm 4.8$& 75.1 $\pm 3.4$&80.6 $\pm 0.2$&88.7 $\pm 0.3$& 87.9 $\pm 1.2$ & 90.7 $\pm 0.6$ &92.3 $\pm 0.5$\\
        FactorGNN & 98.3 $\pm 0.3$ & 96.9 $\pm 0.4$ & 97.6 $\pm 0.4$ &58.7 $\pm 4.1$&68.8 $\pm 9.0$&87.2 $\pm 0.7$&89.8 $\pm 0.6$& 91.5 $\pm 0.7$ & 92.3 $\pm 0.4$ & 90.5 $\pm 1.0$ \\
        \midrule
        {\method} &\textbf{99.4} $\pm 0.0$ & \textbf{98.3} $\pm 0.1$ &  \textbf{99.1} $\pm 0.1$ &\textbf{80.7} $\pm 4.0$&\textbf{84.4} $\pm 1.9$&\textbf{93.7} $\pm 0.1$&\textbf{96.2} $\pm 0.1$& \textbf{97.3} $\pm 0.1$ & \textbf{97.5} $\pm 0.1$ &\textbf{95.0} $\pm 0.3$\\
        \bottomrule
    \end{tabular}
     }
     \vskip -1em
\end{table*}

\subsubsection{Baselines} We include classical heuristics, GNN-based link prediction method, GNN-based representation learning methods for link prediction, disentangled GNNs and GNNs for heterophilic graphs as our baselines. For baselines without the link prediction module, they are performed as encoders in the VGAE manner. Here is a brief introduction of each baseline:
\begin{itemize}[leftmargin=*]
    \item \textbf{CN}~\cite{newman2001clustering}: Common-neighbor index is a classical link prediction method based on connectivity, which assumes that two nodes are more likely to connect if they have many common neighbors.
    \item \textbf{AA}~\cite{adamic2003friends}: Adamic–Adar index is a second-order heuristic based on the assumption that a shared neighbor with large degree is less significant to the measure of a link.
    \item \textbf{VGAE}~\cite{alemi2016deep}: Variational Graph Autoencoder is a generative model for graph representation.  We use a GCN as the encoder where the second layer has two channels for mean and deviations to sample the latent embeddings and a link reconstruction module as the decoder.
    \item \textbf{ARGVA}~\cite{pan2018adversarially}: Adversarially Regularized Variational Graph Autoencoder  is inspired by the adversarial learning methods and the latent representation is enforced to match a prior distribution via an adversarial training scheme.
    \item \textbf{GPR-GNN}~\cite{chien2020adaptive}: Generalized PageRank GNN adaptively learns the GPR weights to optimize the node representation, regardless of the homophilic graph or heterophilic graph.
    \item \textbf{FAGCN}~\cite{bo2021beyond}: Frequency Adaptation Graph Convolutional Networks utilize a selfgating mechanism to adaptively integrate different signals in the process of message passing.
    \item \textbf{LINKX}~\cite{lim2021large}: Linkx decoupled the feature representation and structure representation and use a simple framework to process the decoupled information.
    \item \textbf{GAT}~\cite{velivckovic2017graph}: GAT is a message-passing graph neural network utilizing attention mechanism for better aggregation of neighborhood.
    \item \textbf{DisenGCN}~\cite{ma2019disentangled}: DisenGCN learns disentangled node representations and then utilizes a novel neighborhood routing mechanism to find latent factors that shape the relationship between the nodes. Then for each factor the nodes aggregate the information and get different representations.
    \item \textbf{FactorGNN}~\cite{yang2020factorizable}: FactorGNN shares the similar motivation of disentangling graphs with respect to latent factors. And these models focus more on graph-level tasks.
    \item \textbf{SEAL}~\cite{zhang2018link}: SEAL is a link prediction method that uses a GNN to learn heuristics from local subgraphs automatically.
\end{itemize}
\subsubsection{Configurations} We use a 64-bit machine with Nvidia GPU (Tesla V100, 1246MHz , 32 GB memory) to conduct our experiments. We use ADAM optimization algorithm to train the models. For all methods, the learning rate is initialized to 0.001, with weight decay being 5e-4. We train all the models until converging with a maximum training epoch 2000.

\subsubsection{Evaluation Metrics} Following existing works~\cite{zhang2018link} in evaluating link prediction, area under the curve (AUC) is used as our evaluation metric to measure the prediction accuracy. 

\subsubsection{Implementation Details}
Following a standard way to evaluate link prediction performance~\cite{lichtenwalter2010new}, for each dataset, we randomly split existing edges for the training set, validation set and testing set according to 85\%, 5\% and 10\% as positive samples. Under the same split proportion, we randomly sample nonexistent edges as negative samples with 5 times the number of positive samples. Positive and negative samples are then combined as our training, validation and testing data. For each dataset, we randomly split it 10 times and report the average results. For {\method}, the temperature parameter $\tau$ v is varied among $\{0.1,1\}$ and the teleport parameter $\beta$ is varied from 0.1 to 0.9. The number of factors is varied from 2 to 10 and we set the dimension of representations for each factor as 32. 

\subsection{Link Prediction Performance}

\begin{table}[t]
    \centering
    \small
    \caption{Link Prediction performance (AUC(\%) $\pm$ Std.) on homophilic graphs (``-'' denotes out of memory.)} \label{tab:homo}
    \vskip -1em
    \begin{tabular}{lccc}
    \toprule
    Method & Cora & Citeseer & Photo \\
    \midrule
        AA & 75.9 $\pm 1.5$ & 69.8 $\pm 2.2$& 97.4 $\pm 0.4$\\
        CN & 75.7 $\pm 1.3$ &69.7$\pm 2.3$&97.1 $\pm 0.4$\\
        SEAL &90.4 $\pm 1.1$& 97.0 $\pm 0.5$&-\\
        VGAE & 96.6 $\pm 0.2$ & 97.3 $\pm 0.2$ & 94.9 $\pm 0.8$ \\
        ARGVA & 92.7 $\pm 1.3$ & 94.8 $\pm 0.4$& 89.7 $\pm 2.3$ \\
        GPR-GNN & 94.8 $\pm 0.3$ & 96.0 $\pm 0.3$ & 97.0 $\pm 0.2$\\
        FAGCN & 91.8 $\pm 3.4$ & 89.2 $\pm 5.6$& 94.0 $\pm 1.9$\\
        LINKX & 93.4 $\pm 0.3$& 93.5 $\pm 0.5$& 97.0 $\pm 0.2$\\
        GAT & 97.0 $\pm 0.2$ &  \textbf{98.3} $\pm 0.2$& 97.3 $\pm 0.3$\\
        DisenGCN & 96.6 $\pm 0.3$ & 96.8 $\pm 0.2$& 95.2 $\pm 1.3$\\
        FactorGNN & 92.3 $\pm 1.4$ & 87.8 $\pm 3.6$ & 97.2 $\pm 0.1$\\
        \midrule
        {\method} & \textbf{97.1} $\pm 0.4$ &  \textbf{98.3} $\pm 0.3$& \textbf{97.9} $\pm 0.1$ \\
        \bottomrule
    \end{tabular}
`     
    
    \vskip -1em
\end{table}
To answer \textbf{RQ1}, we evaluate the link prediction performance of {\method} with a comparison on 10 heterophilic and 3 homophilic graphs. Each experiment is repeated 5 times to alleviate randomness. The average link prediction performance measured by AUC with standard deviations on heterophilic graphs and homophilic graphs are shown in Table~\ref{tab:heter} and Table~\ref{tab:homo}, respectively. From the tables, we have the following observations: 
\begin{itemize}[leftmargin=*]
\item Compared with classical heuristics AA and CN, our method outperforms them in all datasets. For Wikipedia, FB100 and Photo datasets with higher average node degree, the performance gap is relatively small. While AA and CN only rely on the graph structural feature, for the other datasets with smaller average node degrees, {\method} is much better due to the disentangled learning of both node features and the graph structure.
\item {\method} has comparable performance on the Chameleon and outperforms other datasets compared with SEAL. It needs to be noticed that SEAL fails to be implemented on graphs with a large number of edges which will lead to the huge memory consumption of the enclosing subgraph extraction before learning, while {\method} is implemented in an end-to-end fashion.
\item Compared with representation learning methods VGAE and ARGVA, {\method} performs better and is much more robust across datasets as there is a larger performance gap between similar datasets in VGAE (DE and ENGB, Amberst41 and Johns Hopkins55). For ARGVA, the adversarial training does not fit well for the link prediction task. Compared with the GCN encoder in them, the modeling of edge factors in {\method} disassembles the complex edge relationships in heterophilic graphs for better structural information mining.
\item For state-of-the-art GNNs targeted for heterophilic graphs GPR-GNN,FAGCN and LINKX, performing as an encoder in VGAE, perform worse in heterophilic datasets, e.g. crocodile, ENGB, compared with vanilla VGAE. Although these methods can successfully improve the node feature learning with respect to the node classification task for heterophilic graphs, they fail to capture different latent relations underlying graph structure using the same set of learned representations. {\method} can better preserve different relational information via separated link reconstruction for different factors and thus outperform these heterophilic graph targeted methods.
\item {\method} also performs better than other disentangled GNN methods and GAT in link prediction. For disentangled GNN models DisenGCN and FactorGNN, they perform factor-wise message-passing with the same graph structure. Although node representations are weighted differently in each factor, the unrelated nodes in each factor will still add unnecessary signals when aggregated with the nodes pertinent to that factor. In contrast, {\method} performs individual message-passing with disjoint adjacency matrices and merges the learned structural information by the proposed disentangled link reconstruction at the end. GAT has good performance and even performs better on Citeseer than {\method} thanks to its multi-head attention mechanism.
\item In general, {\method} can be adapted well both on small and large graph datasets. For Texas and Wisconsin, {\method} shows an apparent improvement of around 10\% compared with baselines. For large graph dataset Arxiv-Year, besides better accuracy, {\method} is more stable with smaller standard deviations. 
\end{itemize}
\begin{figure}
    \centering
    \includegraphics[width=0.8\linewidth]{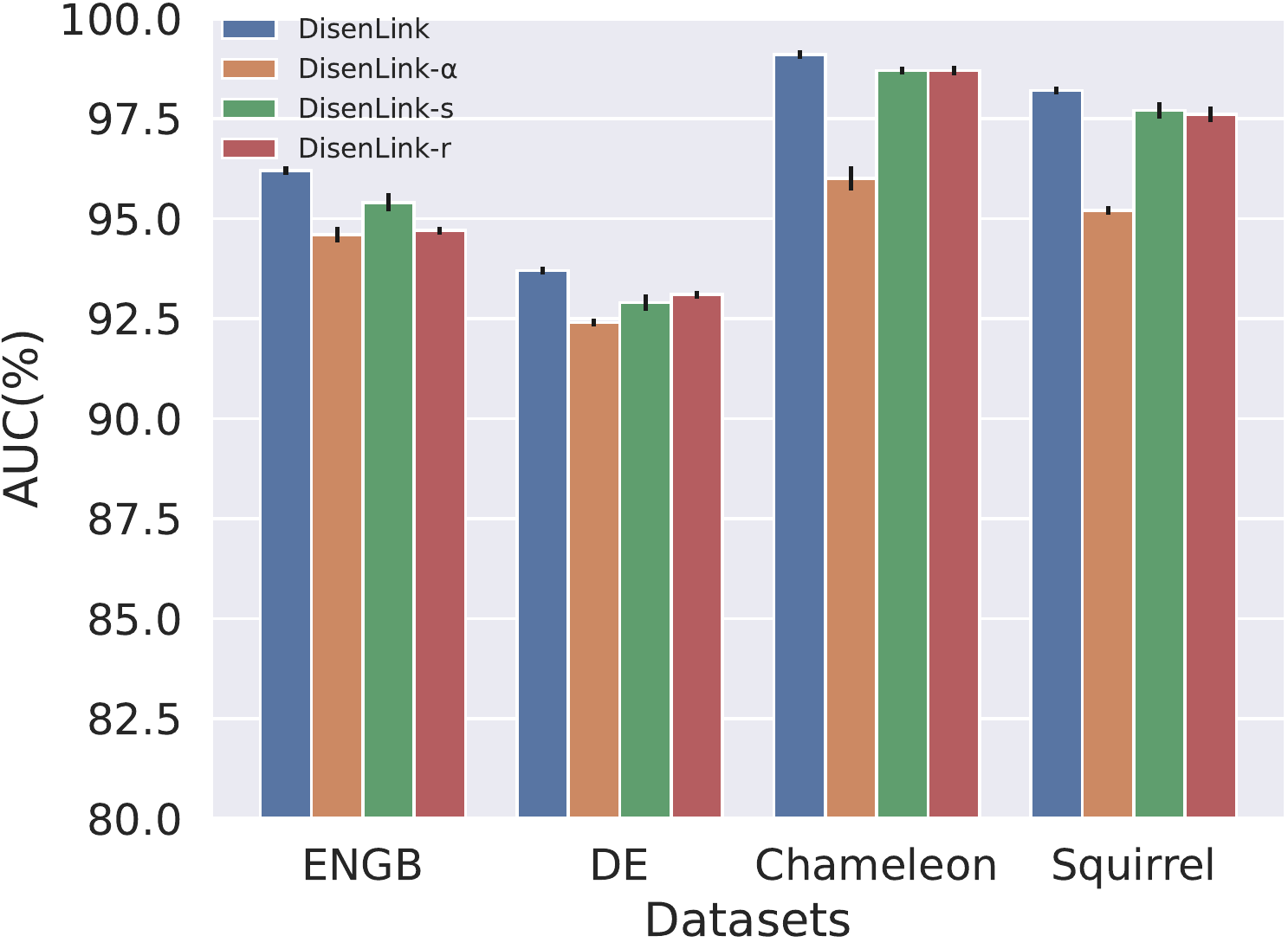}
    \vskip-1.5em
    \caption{Affect of different components in {\method}} \label{fig:ablation}
    \vskip -1em
\end{figure}

\subsection{Ablation Study}
In this subsection, to answer \textbf{RQ2}, we conduct experiments to evaluate the effect of each component of {\method}. Each ablation experiment is conducted 5 times to alleviate the randomness. In particular, we investigate 3 variants of {\method}: {\methoda}, {\methods} and {\methodr}. {\methoda} is formulated by removing all the importance/similarity $\alpha$ involved in {\method} to check the effect of factor-wise importance values. {\methods} is the variant without the proposed factor-aware neighbor selection, i.e. the neighbor of node s $\mathcal{N}_{k}(v_s)$ for each factor in {\methods} becomes same as $\mathcal{N}(v_s)$ to verify the role of factor-aware neighbor selection. {\methodr} is the variant utilizing the vanilla manner of link reconstruction following VGAE~{\cite{kipf2016variational}} as Eq.~{\ref{eq:link_prediciton_equally}} to analyze the influence of disentangled link reconstruction.

\begin{figure}
\centering
\subfigure[ENGB]{
    \includegraphics[width=0.50\linewidth]{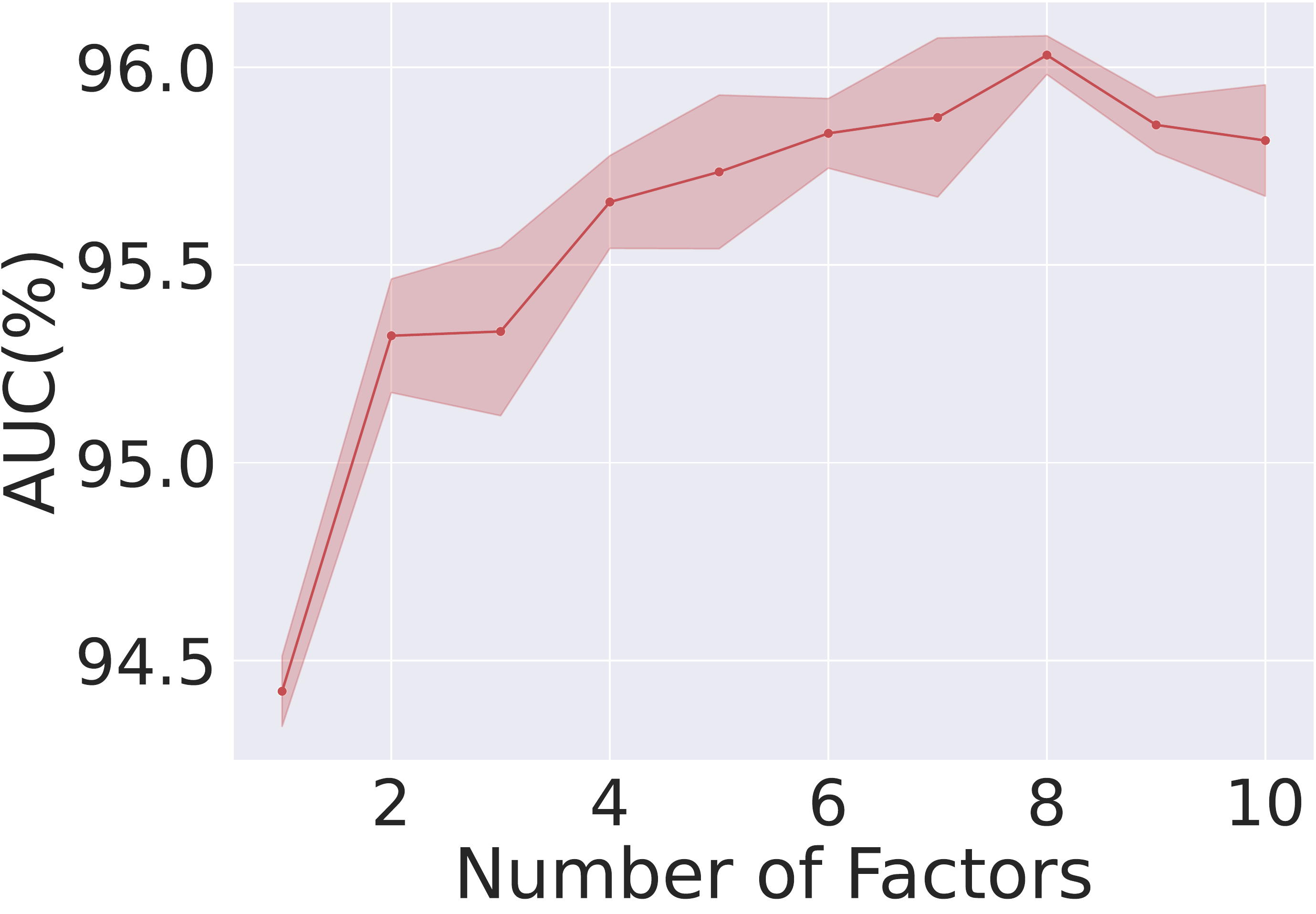}
    \label{fig:engb_nfactor}
}\subfigure[Cora]{
    \includegraphics[width=0.50\linewidth]{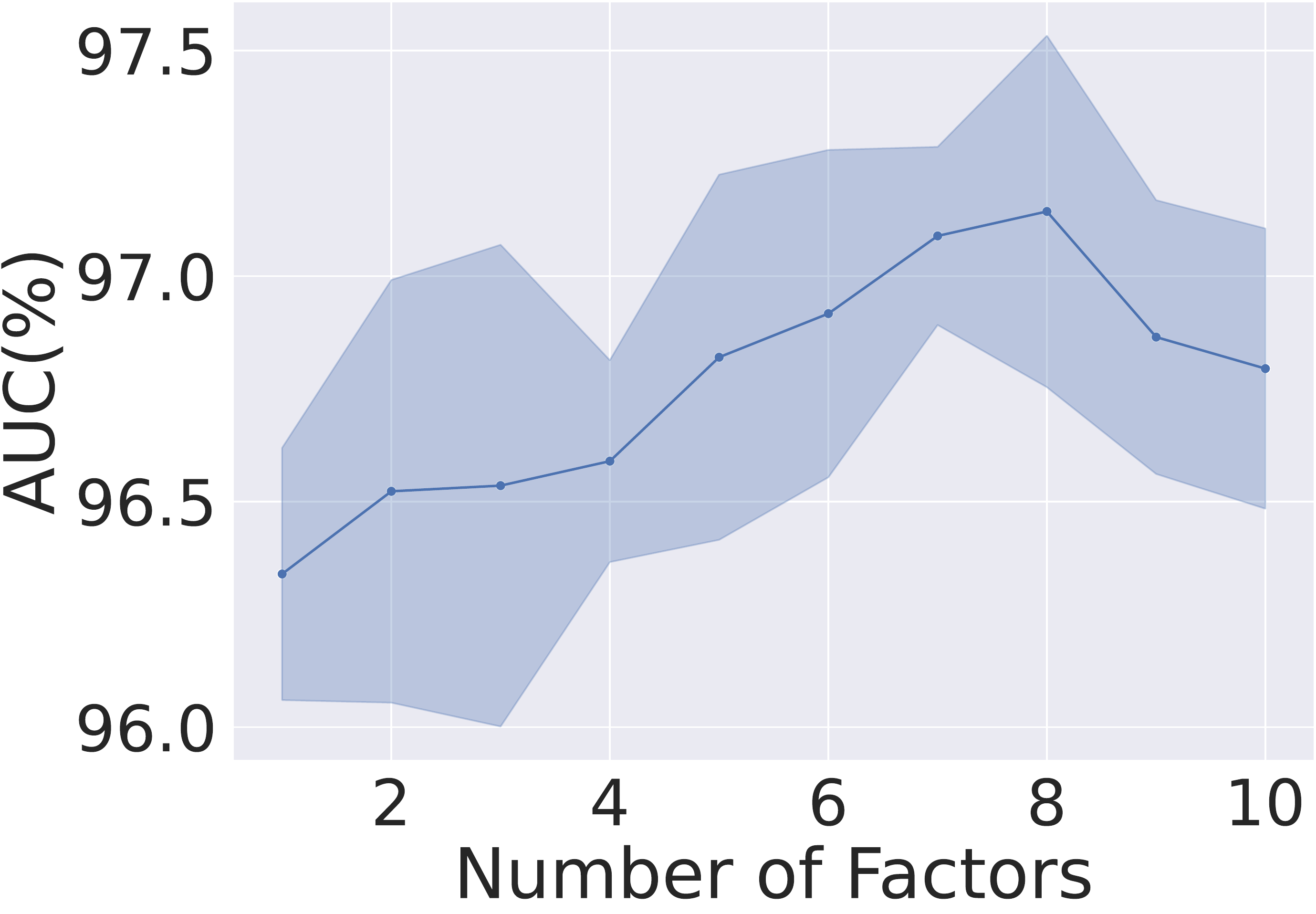}
    \label{fig:cora_nfactor}
}
\vskip -1.5em
\caption{Sensitivity of {\method} on $K$ for Link Prediction }
\vskip -1em
\label{fig:nfactor}
\end{figure}

\textbf{Importance of the disentangled manner} We first examine the advantage of the disentangled learning manner compared with the normal model with a single factor. As shown in the Fig.~\ref{fig:nfactor}, {\method} with multiple factors performs much better than the {\method} with a single factor, especially in the case of heterophilic graph ENGB. It proves that the disentangled representation can facilitate more effective edge relation discovery compared with normal learning. 

\textbf{Significance of the factor-wise importance value} Secondly, the importance value $\alpha$ computed from Eq.~(\ref{eq:attention}) for message passing is crucial for link prediction on heterophilic graphs. We can first compare the performance of 1-factor {\method} reflected from the first data point ($K=1$) in Fig.~{\ref{fig:engb_nfactor}} with VGAE on ENGB dataset. In this case, the main difference between these two models is that 1-factor {\method} takes into the importance value and 1-factor {\method} improves the performance by around 4.1\% compared with VGAE. This is because the $\alpha$ weights node representations in the message-passing process to leverage the negative effect from heterophilic connections. 
And the performance of {\methoda} shown in Fig.~{\ref{fig:ablation}} is much lower compared with {\method} in 4 reported datasets. It demonstrates the importance of relational weighting in message-passing and link reconstruction for link prediction.  

\textbf{Influence of the factor-aware neighbor selection} We also conduct ablation studies to explore the influence of the factor-aware neighbor selection on the link prediction on heterophilic graphs. In Fig.~{\ref{fig:ablation}}, {\method} outperforms {\methods} which reflects the selection of the most important factor between two nodes to formulate $K$ disjoint neighborhoods is very helpful for link prediction.

\textbf{Influence from link reconstruction manner of {\method}} In {\method}, the link reconstruction utilizing the learnt disentangled representation is implemented as Eq.~{\ref{eq:re}}. Here, we compare it with model variant {\methodr}. The performance difference between them shown in the Fig.~{\ref{fig:ablation}} verifies the effectiveness of our disentangled link reconstruction which assigns different factors with different importance during link prediction.

\begin{figure}
\centering
\vskip -1em
\subfigure[ENGB]{
    \includegraphics[width=0.50\linewidth]{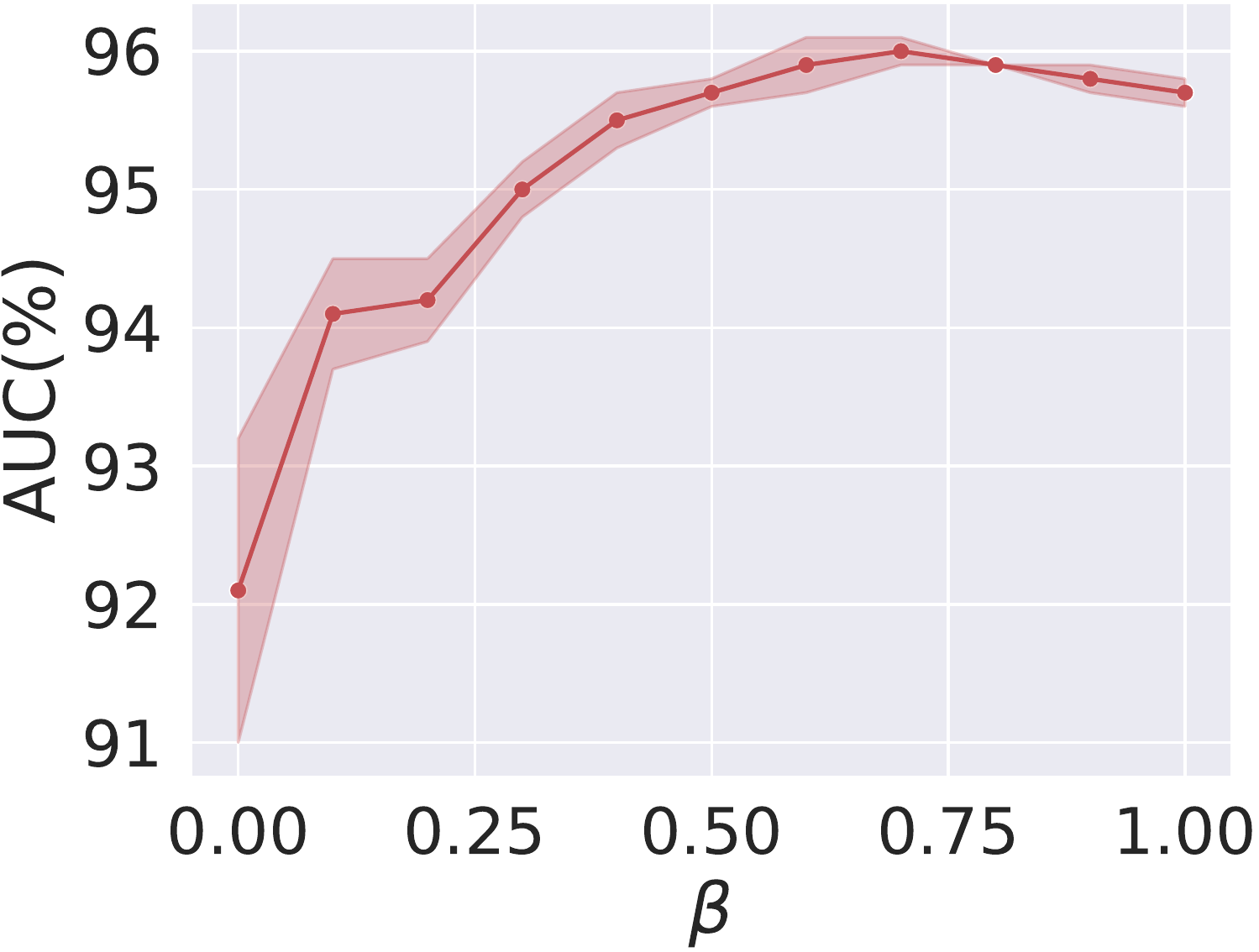}
    \label{fig:engb_beta}
}\subfigure[ArXiv-year]{
    \includegraphics[width=0.50\linewidth]{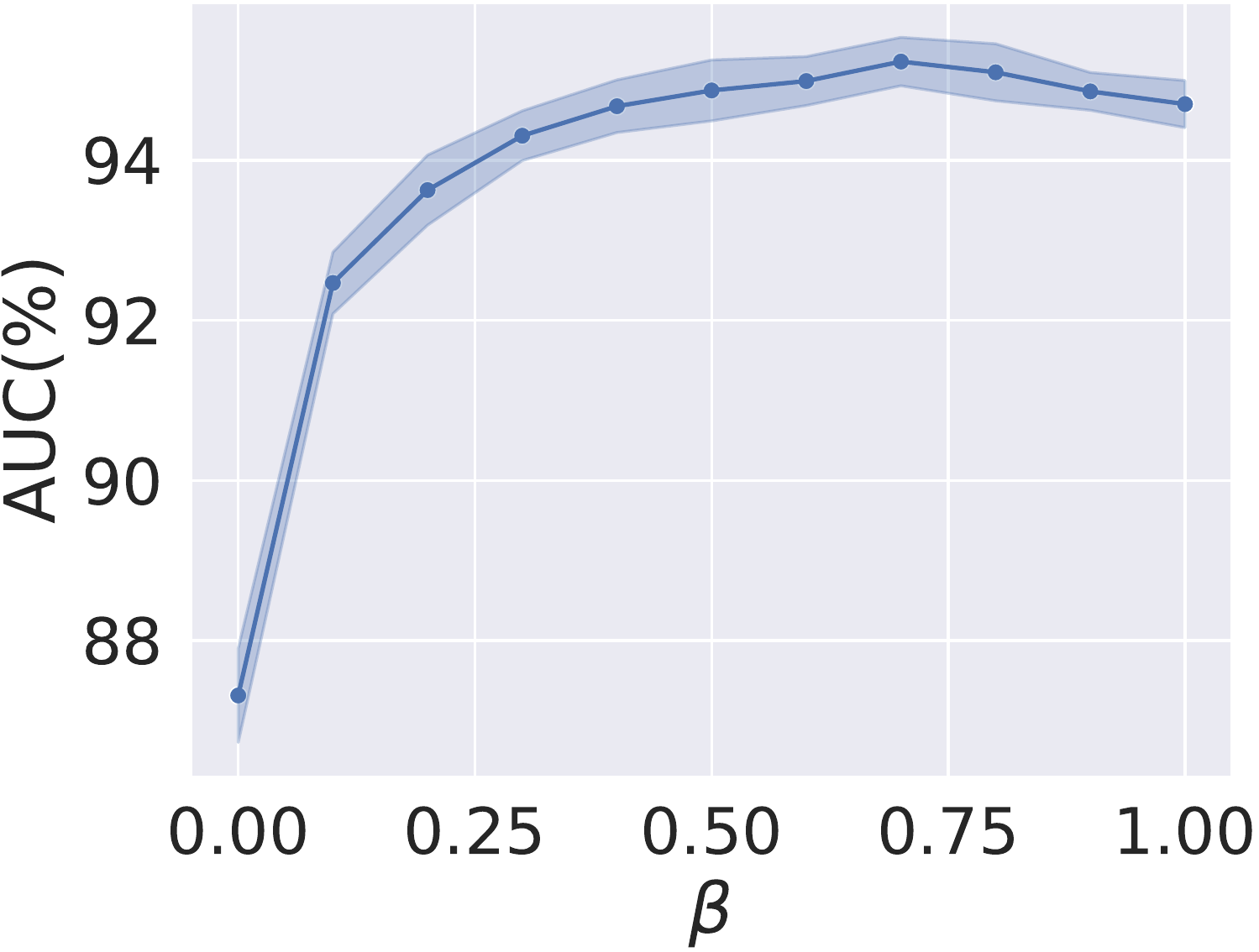}
    \label{fig:year_beta}
}\vskip -2em
\caption{Parameter Sensitivity of $\beta$ }
\label{fig:beta}
\vskip -1.5em
\end{figure}
\subsection{Hyperparameter Sensitivity Analysis}
In this section, we investigate the sensitivity of {\method} on two important hyper-parameters, i.e., the number of factors $K$ and the $\beta$ in message passing. To understand the sensitivity on $K$, we fix $\beta$ as 0.5 and vary $K$ from 1 to 10. The dimension of latent representation is 32 for each factor. We conduct the experiments on one homophilic graph Cora and one heterophilic graph ENGB as shown in Fig.~\ref{fig:nfactor}. According to the Fig.~\ref{fig:cora_nfactor}, when the number of factors is equal to 8 which is slightly bigger than its actual number of class 7, experiments on Cora reach the best performance. For ENGB which has nodes with 2 classes, the performance reaches the peak around $k=6$ and fluctuates slightly after it is shown in Fig.~\ref{fig:engb_nfactor}. This is because the number of ground truth can not reflect the actual number of latent factors in heterophilic graphs due to the complex intra-class distribution. 

To understand the sensitivity of $\beta$, we fix $K$ as 5 and vary $beta$ from 0 to 1. We run the experiment 5 times and the link prediction performance on Cora, ENGB and ArXiv-year are shown in Fig.~\ref{fig:beta}. The sensitivity on $\beta$ shown in Fig.~\ref{fig:beta} reflects how the proportion of the central node content in factor-wise message-passing affects the link prediction. The existence of $\beta$ can help the preservation of locality and prevent uninformative node representation in each factor since the disjoint factor graph includes isolated nodes. As a result, when $\beta> 0$, the performance will be much improved. And when $\beta$ is near 1, the performance will drop due to the lack of neighbor feature information.
\begin{figure}
\centering     
\subfigure[The representation correlation of {\methods}]
{\label{fig:corre_without}
\includegraphics[width=0.76\linewidth]{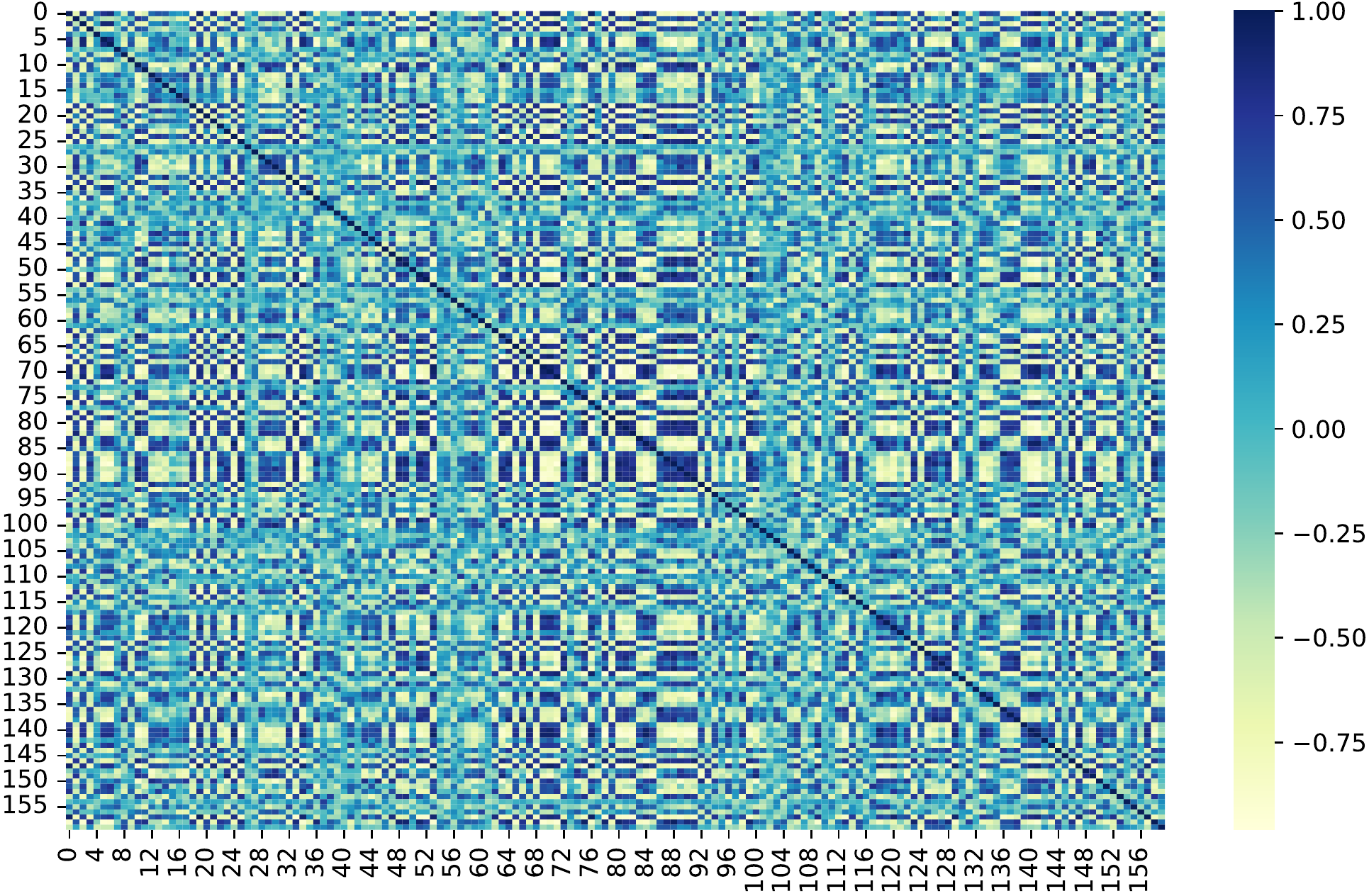}}
\vskip -0.5em
\subfigure[The representation correlation of {\method}]
{\label{fig:corre_with}
\includegraphics[width=0.76\linewidth]{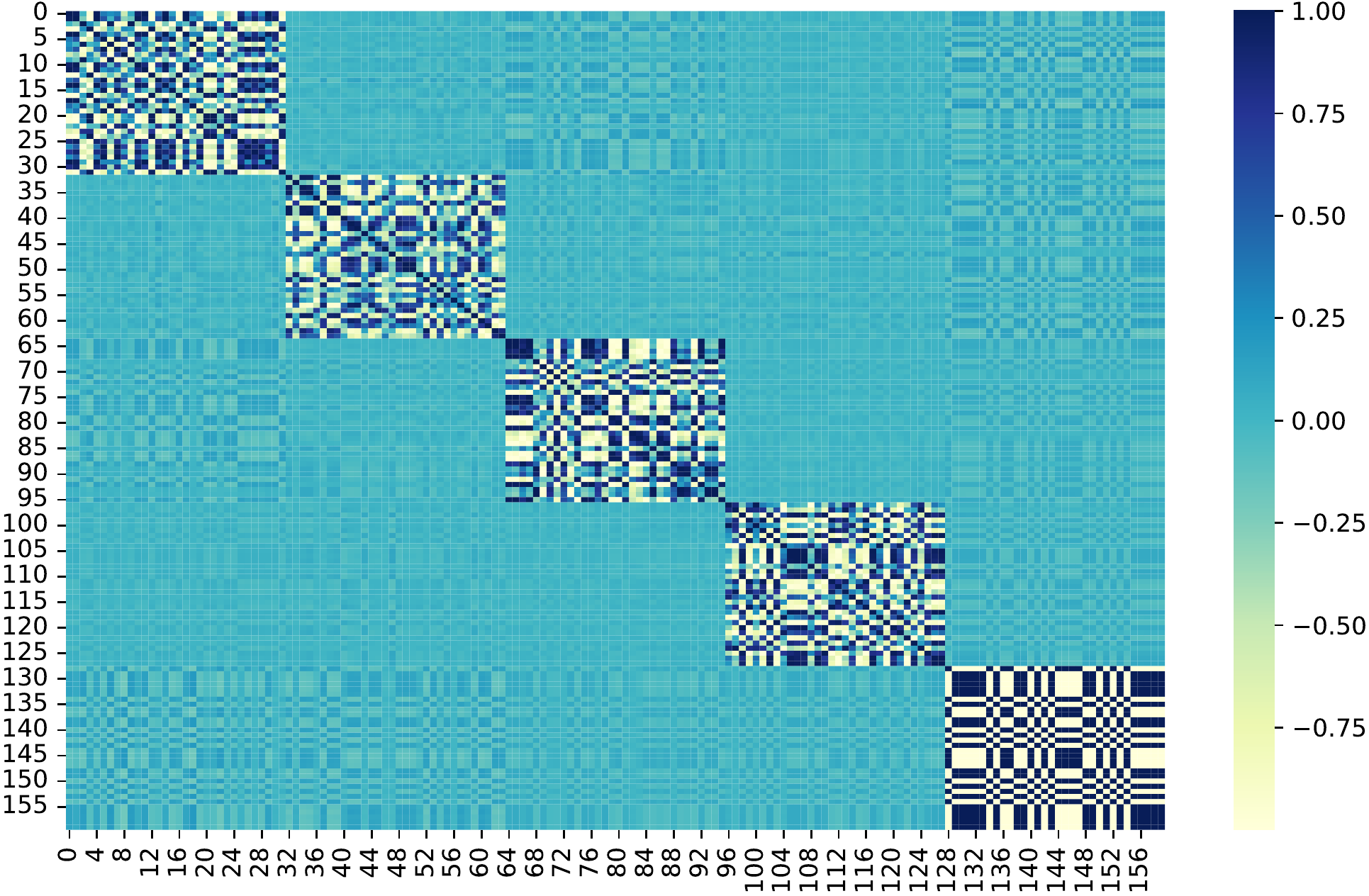}}
\vskip -1.5em
\caption{Disentangled feature correlation coefficient without/with factor-aware neighbor selection}
\label{fig:coeffcient}
\vskip -1em
\end{figure}



\subsection{Analysis on Disentangled Representation}
To further analyze if ${\method}$ can learn high-quality disentangled representation, in this subsection, we compute the absolute value of the correlations between the elements of the disentangled representation learned. Specifically, we train {\method} and get the latent representation of the nodes. Let $\mathbf{H} = [\mathbf{H}_1, \dots, \mathbf{H}_K] \in \mathbb{R}^{N \times K\cdot d}$ be the learned representation of nodes, with the $s$-th row of $\mathbf{H}$ being the disentangled representation of node $v_s$, i.e., $\mathbf{h}_i = [\mathbf{h}_{s,1},\dots,\mathbf{h}_{s,K}]$. $d$ is the dimension of each $\mathbf{h}_{s,k}$. We then calculate the correlation between the column of $\mathbf{H}$ to get the correlation. We set $K=5$ and $d=32$. We conduct this for both {\method} and {\methods}. The absolute value of the correlation on Chameleon is shown in Fig.~{\ref{fig:coeffcient}}. Compared with the mixed correlation in the case of {\methods} shown in Fig.~{\ref{fig:corre_without}}, the correlation plot of {\method} in Fig.~{\ref{fig:corre_with}} shows $5$ explicit blocks in the diagonal. The difference between {\method} and {\methods} implies the factor-aware neighbor selection can produce independent factors and promote the disentanglement of the output representations, which answers \textbf{RQ3}.

\section{CONCLUSION AND FUTURE WORK}
Due to the entanglement of latent factors in none-homophily graphs, we investigate how to perform effective link prediction on heterophilic graphs in this paper and propose a disentangled link prediction framework {\method}, which includes edge factor discovery with selection and factor-wise message passing for better node representations and a disentangled link reconstruction module to benefit the formation of links. Experiments on real-world datasets with different levels of disassortativity demonstrate the effectiveness of {\method} for link prediction on heterophilic graphs. Further experiments are conducted to understand the influence of each component and the sensitivity of hyperparameters in {\method}. 
In the future, several interesting directions are left to be explored. For example, while our work focuses on link prediction, more tasks, such as node clustering, unsupervised node classification, can be extended for our model. Another potential direction is to investigate the learning capacity of {\method} in heterogeneous graphs since heterogeneous graphs naturally have various kinds of nodes and links.
\bibliographystyle{ACM-Reference-Format}
\bibliography{reference}

\end{document}